\documentclass[10pt,twocolumn,letterpaper]{article}

\usepackage{iccv}
\usepackage{times}
\usepackage{epsfig}
\usepackage{graphicx}
\usepackage{amsmath}
\usepackage{amssymb}
\usepackage{ctable}
\usepackage{multirow}
\usepackage{enumitem}
\usepackage{stmaryrd}

\newcommand{\CG}[1]{\textcolor{blue}{[Chuang]}} 
\newcommand{\AT}[1]{\textcolor{red}{[Antonio]}} 
\newcommand{\weichiu}[1]{\textcolor{orange}{[Wei-Chiu]}} 

\usepackage[pagebackref=true,breaklinks=true,letterpaper=true,colorlinks,bookmarks=false]{hyperref}

\iccvfinalcopy 

\def\iccvPaperID{} 

\ificcvfinal\fi
\begin{document}

\title{The Sound of Motions}

\author{Hang Zhao$^{1}$, Chuang Gan$^2$, Wei-Chiu Ma$^1$, Antonio Torralba$^1$ \\
$^1$MIT \quad $^2$MIT-IBM Watson AI Lab \\
{\tt\small \{hangzhao,chuangg,weichium,torralba\}@mit.edu}
}

\maketitle

\begin{abstract}

Sounds originate from object motions and vibrations of surrounding air. 
Inspired by the fact that humans is capable of interpreting sound sources from how objects move visually, we propose a novel system that explicitly captures such motion cues for the task of sound localization and separation.
Our system is composed of an end-to-end learnable model called Deep Dense Trajectory (DDT), and a curriculum learning scheme. It exploits the inherent coherence of audio-visual signals from a large quantities of unlabeled videos.
Quantitative and qualitative evaluations show that comparing to previous models that rely on visual appearance cues, our motion based system improves performance in separating musical instrument sounds. Furthermore, it separates sound components from duets of the same category of instruments, a challenging problem that has not been addressed before.

\end{abstract}

\section{Introduction}
In a scorching afternoon, you relax under the shadow of a tree and enjoy the breeze. You notice that the tree branches are \emph{vibrating} and you hear a \emph{rustling sound}.  Without a second thought, you realize that the sound is caused by the leaves \emph{rubbing} one another. Despite a short notice, humans have the remarkable ability to connect and integrate signals from different modalities and perceptual inputs. In fact, the interplay among senses are one of the most ancient scheme of sensory organization in human brains \cite{stein1993merging} and is the key to understand the complex interaction of the physical world.

\begin{figure}[t]
	\centering
	\includegraphics[width=0.85\linewidth]{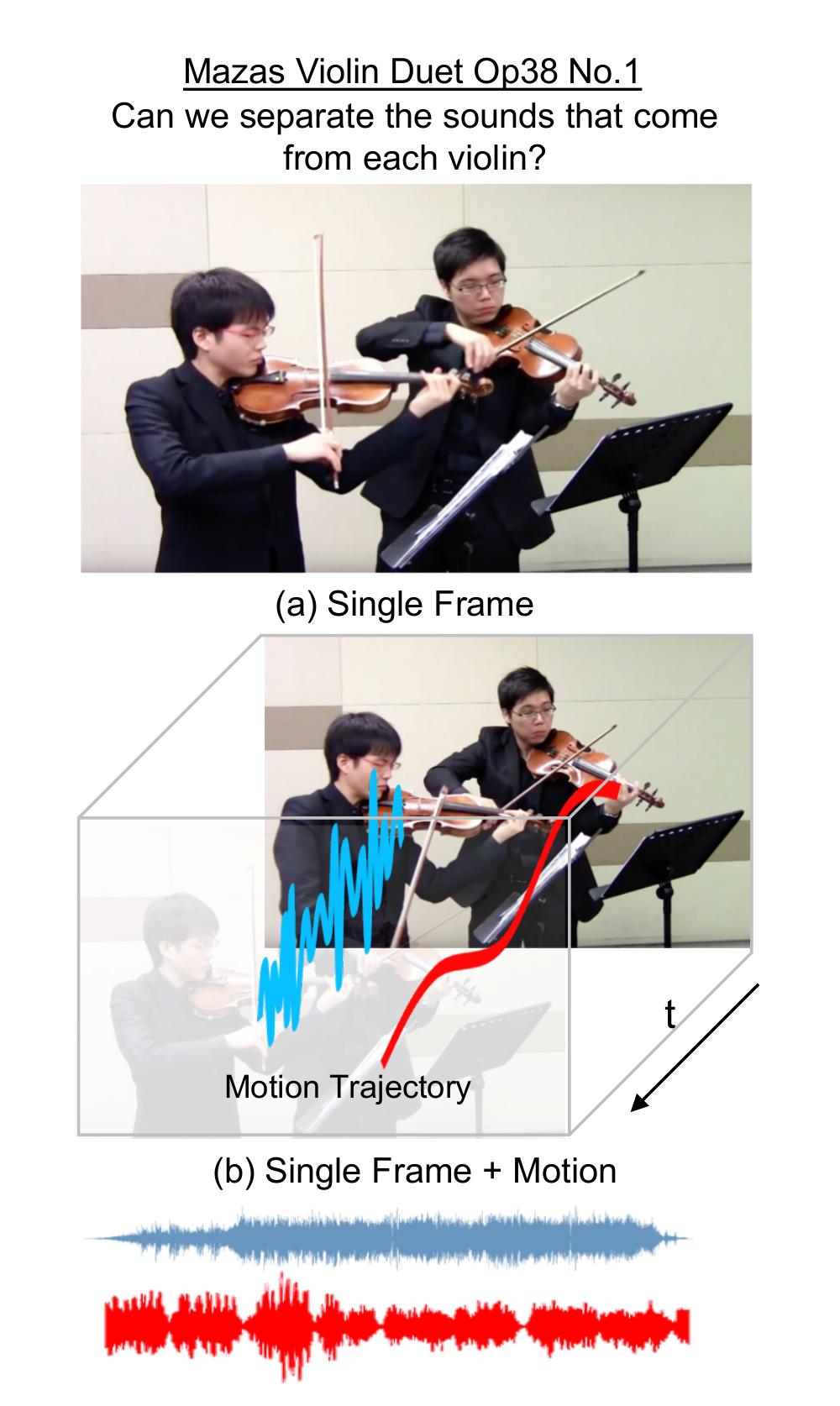}
	\caption[]{Motion matters: When watching a violin duet video, we can separate the melody from harmony. (a) Yet it is hard to tell the sources without looking or with only one glance. (b) By watching for a bit longer, we can differentiate who is playing the first violin and who is playing the second by associating their motions with the tempo of the music.
		In this work, we take inspirations from human to disambiguate and separate the sounds from multiple sources by exploring motion cues.
	}
	\label{fig:teaser}
\end{figure}

With such inspiration in mind, researchers have been painstakingly developing models that can effectively exploit signals from different modalities. 
Take audio-visual learning for example, various approaches have been proposed such as sound recognition \cite{aytar2016soundnet,arandjelovic2017look,korbar2018co}, sound localization \cite{Hershey1999,Kidron2005,Ngiam2011,arandjelovic2017objects,ephrat2018looking}, \etc. In this work, we are particularly interested in the task of sound source separation \cite{ephrat2018looking,Zhao_2018_ECCV,gao2018object-sounds}, where the goal is to distinguish the components of the sound and associate them with the corresponding objects.
While current source separation methods achieve decent results on respective tasks, they often ignore the motion cues and simply rely on the static visual information. Motion signals, however, are of crucial importance for audio-visual learning, in particular when the objects making sounds are visually similar. Consider a case where two people are playing violin duets, as depicted in Figure \ref{fig:teaser}. It is virtually impossible for humans to separate their melody from harmony by peaking at a single image. Yet if we see the movement of each person for a while, we can probably conjecture according to the temporal repetition of the motions and the beats of music. This illustration serves to highlight the importance of motion cues in the complex multi-modal reasoning. Our goal is to mimic, computationally, the ability to reason about the synergy between visual, audio, and motion signals\footnote{We encourage the readers to watch the video \url{https://www.youtube.com/watch?v=XDuKWUYfA_U} to get a better sense of the difficulty of this task.}.

We build our model upon previous success of Zhao~\etal~\cite{Zhao_2018_ECCV}. Instead of relying on image semantics, we explicitly consider the temporal motion information in the video. In particular, we propose an end-to-end learnable network architecture called Deep Dense Trajectory (DDT) to learn the motion cues necessary for the audio-visual sound separation. As the interplay among different modalities are very complex, we further develop a curriculum learning scheme. By starting from different instruments and then moving towards same types, we force the model to exploit motion cues for differentiation.

We demonstrate the effectiveness of our model on two recently proposed musical instrument datasets, MUSIC~\cite{Zhao_2018_ECCV} and URMP~\cite{li2019creating}.
Experiments show that by explicitly modeling the motion information, our approach improves prior art on the task of audio-visual sound source separation. More importantly, our model is able to handle extremely challenging scenarios, such as duets of the same instruments, where previous approaches failed significantly.



\section{Related Work}
\label{sec:related}


\paragraph{Sound source separation.}

Sound source separation is a challenging classic problem, and is known as the ``cocktail party problem"~\cite{mcdermott2009cocktail,haykin2005cocktail} in the speech area. Algorithms based on Non-negative Matrix Factorization (NMF) \cite{virtanen2007monaural,cichocki2009nonnegative,smaragdis2003non} were the major solutions to this problem. More recently, several deep learning methods have been proposed, where Wang \etal gave an overview~\cite{wang2017supervised} on this series of approaches. Simpson~\etal~\cite{simpson2015deep} and Chandna~\etal~\cite{chandna2017monoaural} used CNNs to predict time-frequency masks for music source separation and enhancement. To solve the identity permutation problem in speech separation, Hershey~\etal~\cite{hershey2016deep} proposed a deep learning-based clustering method, and Yu \etal~\cite{yu2017permutation} proposed a speaker-independent training scheme. While these solutions are inspiring, our setting is different from the previous ones in that we use additional visual signals to help with sound source separation.

\paragraph{Audio-visual learning.}
Learning the correspondences between vision and sound has become a popular topic recently. 
One line of work has explored representation learning from audio-visual training. Owens \etal~\cite{owens2016ambient} used sound signals as supervision for vision model training; Aytar \etal~\cite{aytar2016soundnet} used vision as supervision for sound models; Arandjelovic \etal~\cite{arandjelovic2017look} and Korbar \etal~\cite{korbar2018co} trained vision and sound models jointly and achieve superior results.
Another line of work explored sound localization in the visual input~\cite{izadinia2013multimodal,Hershey1999,arandjelovic2017objects,senocak2018learning,Zhao_2018_ECCV}. More recently, researchers used voices and faces to do biometric matching \cite{nagrani2018seeing}, generated sounds for videos \cite{zhou2017visual}, generated talking faces~\cite{zhou2019talking}, and predicted stereo sounds~\cite{gao20182} or 360 ambisonics~\cite{morgado2018self} from videos.

Although a few recent papers have demonstrated how visual cues could help with music separation \cite{Zhao_2018_ECCV,gao2018object-sounds}, their visual cues mostly come from appearance, which can be obtained from a single video frame. Our work differentiate from those in that we explicitly model motion cues, to make good use of the video input.

\paragraph{Sounds and motions.}

Early works in vision and audition have explored the strong relations between sounds and motions.
Fisher~\etal~\cite{fisher2001learning} used a maximal mutual information approach and Kidron~\etal~\cite{Kidron2005,izadinia2013multimodal} proposed variations of canonical correlation methods to discover such relations. 

Lip motion is a useful cue in the speech processing domain, Gabbay \etal~\cite{gabbay2017seeing} used it for speech denoising; Chung \etal~\cite{chung2017lip} demonstrated lip reading from face videos. Ephrat~\etal~\cite{ephrat2018looking} and Owens~\etal~\cite{owens2018audio} demonstrated speech separation and enhancement from videos. 

The most related work to ours is \cite{barzelay2007harmony}, which claimed the tight associations between audio and visual onset signals, and use the signals to perform audio-visual sound attribution. In this work, we generalize their idea by learning an aligned audio-visual representations for sound separation.

\begin{figure*}[ht]
	\centering
	\includegraphics[width=1.00\textwidth]{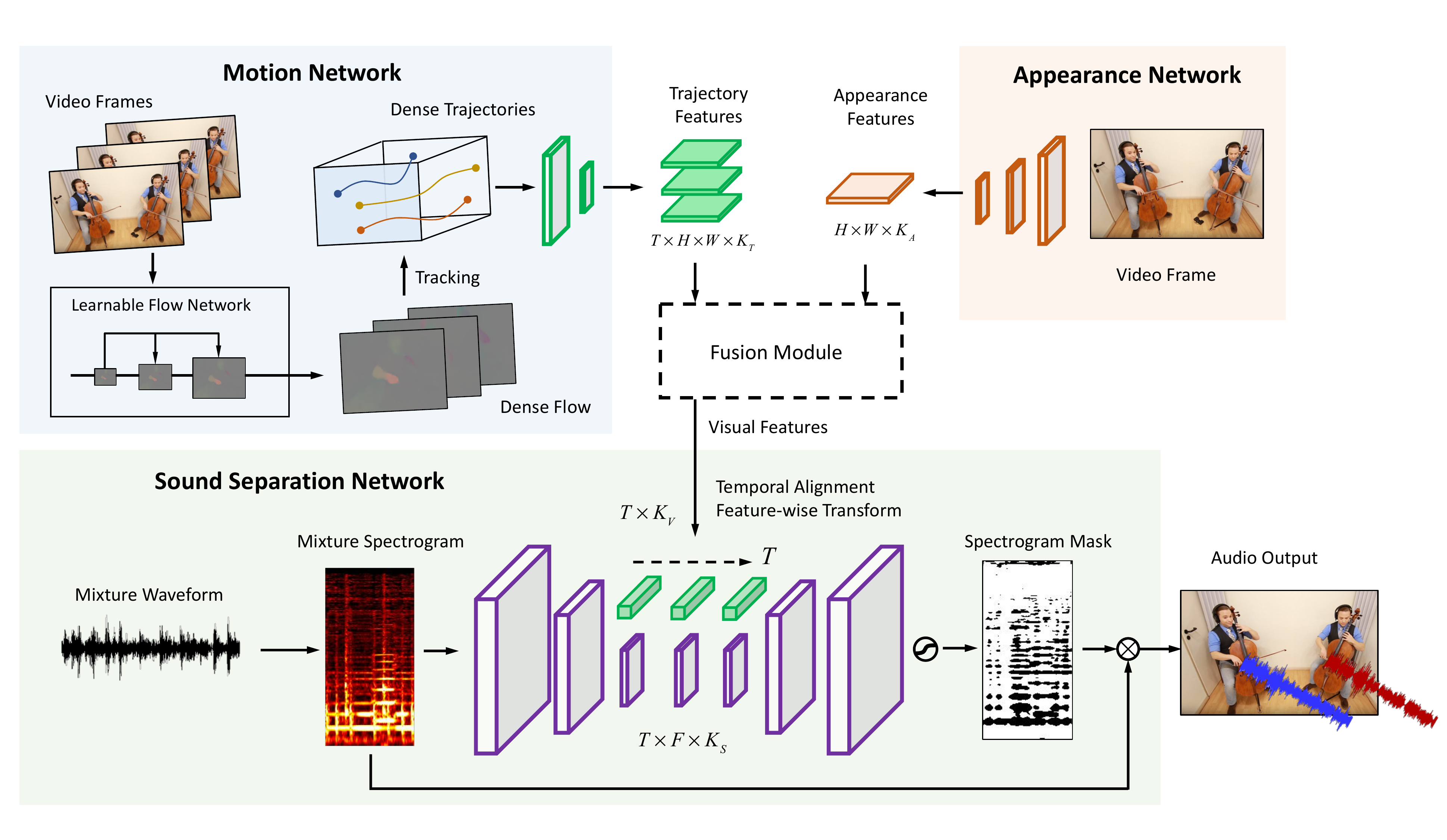}
	\caption{An overview of model architecture. Our framework is consist of four components: a motion network, an appearance network, a fusion module, and a sound separation network. The motion network takes a sequence of frames and outputs trajectory features; appearance network takes the first video frame and outputs appearance features; fusion module fuses appearance and trajectory features; sound separation network separates the input audio conditioned on the visual features.}
	\label{fig:model}
\end{figure*}

\paragraph{Motion representation for videos.}
Our work is in part related to motion representation learning for videos, as we are working on videos of actions. Traditional techniques mainly use handcrafted spatio-temporal features, like space-time interest points~\cite{laptev2005space}, HOG3D~\cite{klaser2008spatio}, dense trajectories~\cite{wang2011action}, improved dense trajectories~\cite{wang2013action} as the motion representations of videos.  Recently, works have shifted to learning representations using deep neural networks. There are three kinds of successful architectures to capture motion and temporal information in videos: (1) two-stream CNNs~\cite{simonyan2014two}, where motion information is modeled by taking optical flow frames as network inputs; (2) 3D CNNs~\cite{tran2015learning}, which performs 3D convolutions over the spatio-temporal video volume; (3) 2D CNNs with temporal models on top such as LSTM~\cite{donahue2015long}, Attention~\cite{long2018attention,bian2017revisiting}, Graph CNNs~\cite{wang2018videos}, \etc. More recently, researchers proposed to learn motion/trajectory representations for action recognition~\cite{fan2018end,zhao2018trajectory,gan2018geometry}.
In contrast to action recognition or localization, our goal is to find correspondence between sound components and movements in videos.

\section{Approach}
\label{sec:approach}
In this section, we first introduce the mix-and-separate framework we used for the audio-visual sound separation. Then we present the model architecture we used for learning motion representations for audio-visual scene analysis. Finally, we introduce the curriculum training strategy for better sound separation results. 

\subsection{Mix-and-Separate for Self-supervised Learning}
Our approach adopted the Mix-and-Separate framework~\cite{Zhao_2018_ECCV} for vision guided sound separation.
Mixture and separated audio ground truths are obtained by mixing the the audio signals from different video clips.
And then the task of our model is to separate the audio tracks from mixture conditioned on their corresponding visual inputs.
Critically, although the neural network is trained in a supervised fashion, it does not require labeled data. Thus the training pipeline can be considered as self-supervised learning.

During training, we randomly select $N$ video clips with paired video frames and audios $\{V_n, S_n\}$, and then mix their audios to form a synthetic mixture $S_{mix} = \sum_{n=1}^N S_n$. Given one of the $N$ video clips, our model $f$ will extract visual features and audio features for source separation $\hat{S_n} = f(S_{mix}, V_n)$. The direct output of our model is a binary mask that will be applied on the input mixture spectrogram, where the ground truth mask of the $n$-th video is determined by whether the target sound is the dominant component in the mixture, 
\begin{equation}
M_n(u, v) = \llbracket S_n(u, v) \ge S_m(u, v)\rrbracket, \quad \forall m=(1,...,N),
\end{equation}
where $(u, v)$ represents the time-frequency coordinates in the spectrogram $S$. The model is trained with per-pixel binary cross-entropy loss.

\subsection{Learning Motions with Deep Dense Trajectories}
\label{sec:approach_ddt}
We use pixel-wise trajectories as our motion features for its demonstrated superior performance in action recognition tasks~\cite{wang2013action}. 

Given a video, the dense optical flow for each frame of the video at time $t$ is denoted as $\omega_t = (u_t, v_t)$,
and we represent the coordinate position of each tracked pixel as $P_t = (x_t, y_t)$. Then the pixels in adjacent frames can associated as $P_{t+1} = (x_{t+1}, y_{t+1}) = (x_t, y_t) + \omega|_{(x_t, y_t)}$, and the full trajectory of a pixel is the concatenation of its coordinates over time $(P_{t}, P_{t+1}, P_{t+2}, ...)$. We use position invariant displacement vectors as the trajectory representation $\mathcal{T} = (\Delta P_{t}, \Delta P_{t+1}, \Delta P_{t+2}, ...)$, where $\Delta P_{t} = (x_{t+1} - x_t, y_{t+1} - y_t)$.

We note that the aforementioned operators are all differentiable, so they can fit into a learnable neural network model. Given the recent advances on CNN-based optical flow estimation, we incorporate a state-of-the-art optical flow model PWC-Net~\cite{sun2018pwc} into our system. So our whole system is an end-to-end learnable pixel tracking model, we refer to it as Deep Dense Trajectory network (DDT).

In previous works on trajectories~\cite{wang2013action}, people usually sub-sample, smooth and normalize pixel trajectories to get extra robustness. We do not perform these operations since we assume that the dense, noisy signals can be handled by the learning system. To avoid tracking drift, we first perform shot detection on the input untrimmed videos, and then track within each video shot.

\subsection{Model Architectures}
\label{sec:arch}

Our full model is shown in Figure~\ref{fig:model}. It is comprised of four parts: a motion network, an appearance network, a fusion module and a sound separation network. We detail them below.

\paragraph{Motion Network.}
The motion network is designed to capture the motion features in the input video, on which the sound separation outputs are conditioned. We introduce Deep Dense Trajectories (DDT) network here, which is an end-to-end trainable pixel tracking network. The DDT network is composed of three steps:
\begin{enumerate}[label=(\roman*)] 
	\item Dense optical flow estimation.
	This step enables the followup trajectory estimation, and it can be achieved by an existing CNN-based optical flow network. We choose the state-of-the-art PWC-Net~\cite{sun2018pwc} for its lightweight design and fast speed.
	PWC-Net estimates optical flow at each level in the feature pyramid, then uses the estimated flow to warp the feature at the next level and constructs a cost volume. 
	
	\item Dense trajectory estimation.
	This step takes dense optical flows as input to form dense trajectories. As discussed in Section~\ref{sec:approach_ddt}, 
	the position of each pixel at the next time stamp is estimated as the current position added by the current optical flow field.
	So the whole trajectory is estimated by iteratively tracking the points according to optical flow fields. In our neural network model, this process is implemented as an iterative differentiable grid sampling process. Specifically, we start with a regular 2D grid $G_0$ for the first frame; then for each frame at time $t$, we sample its optical flow field $\omega_t$ according to current grid $G_t$ to estimate the grid at next time stamp, $G_{t+1} = G_{t} + \text{grid\_sample}(\omega_t, G_{t})$. After tracking, our dense trajectories are given by
	\begin{align*}
	\mathcal{T} &= (\Delta P_{0}, ..., \Delta P_{t}, ...) \\
	&= (\text{grid\_sample}(\omega_0, G_{0}), ..., \text{grid\_sample}(\omega_t, G_{t}), ...),
	\end{align*}
	where $t=(1, ..., T)$. The dimension of trajectories $\mathcal{T}$ is $T\times H\times W\times 2$, where the last dimension represents the displacements in $x$ and $y$ direction.
	
	\item Dense trajectory feature extraction. 
	A CNN model is further applied to extract the deep features of these trajectories, the choice of architecture can be arbitrary. Here we use an I3D model, which demonstrated good capability in capturing spatiotemporal features~\cite{carreira2017quo}. It is a compact design which inflates 2D CNN into 3D so that 3D filters can be bootstrapped from pretrained 2D filters.
	
\end{enumerate}



\begin{figure}[t]
	\centering
	\includegraphics[width=1.0\linewidth]{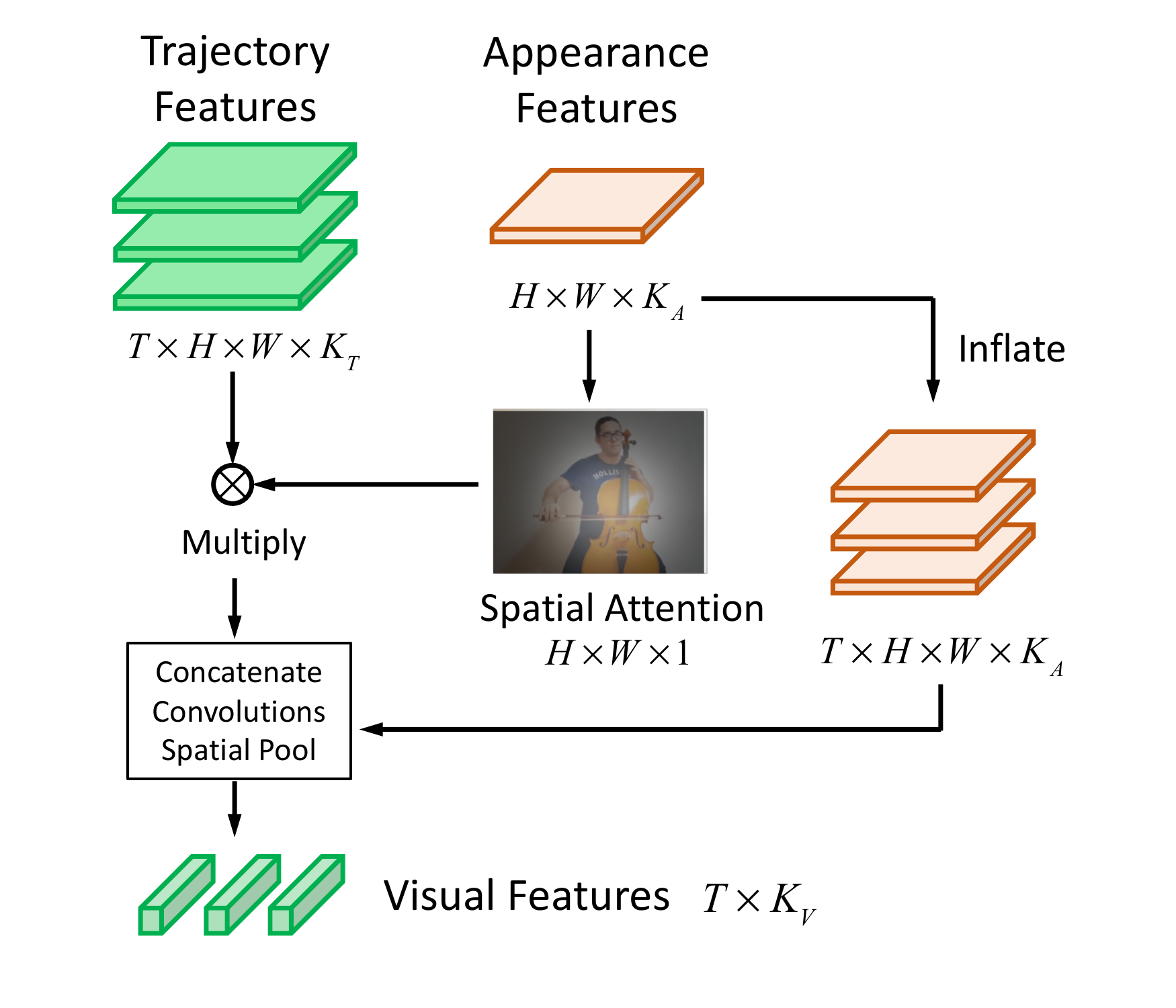}
	\vspace{-0.1in}
	\caption{Fusion module of the model in Figure~\ref{fig:model}. There are two choices: (a) is a concatenation-based fusion where trajectory features and appearance features are stacked together; (b) get a spatial attention map from appearance features, and use it to gate trajectory features. }
	\label{fig:model_fusion}
\end{figure}
\paragraph{Appearance Network}
The appearance network extracts semantic information from the input video. In terms of architecture, we use ResNet-18 \cite{he2016deep} by removing the layers after spatial average pooling. We only take the first frame as input so that the trajectory feature maps are strictly registered with the appearance feature maps. The appearance and trajectory features are then fused to form the final visual features.

\paragraph{Attention based Fusion Module}
To fuse the appearance and trajectory features, we first predict a spatial attention map from the RGB features, and use it to modulate trajectory features. As shown in Figure \ref{fig:model_fusion}, from the appearance feature we predict a single-channel map activated by \texttt{sigmoid}, with size $H\times W \times 1$. It is inflated in time and feature dimension, and multiplied with the trajectory feature from the Motion Network. Then appearance features are also inflated in time, and concatenated with the modulated trajectory features. After a couple of convolution layers, we perform max pooling to obtain the final visual feature. Such attention mechanism helps the model to focus on important trajectories.

\paragraph{Sound Separation Network}
The sound separation network takes in the spectrogram  of sound, which is the 2D time-frequency representation; and predicts a spectrogram mask conditioned on the visual features.
The architecture of sound separation network takes the form of a U-Net \cite{ronneberger2015u}, so that the output mask size is the same as the input. In the middle part of the U-Net, where the feature maps are the smallest, condition signals from visual features are inserted. 
The way to incorporate visual features is by (1) aligning visual and sound features in time; (2) applying Feature-wise Linear Modulation (FiLM)~\cite{perez2017film} on sound features. FiLM refers to a feature-wise affine transformation, formally
\begin{equation}
\text{FiLM}(f_s) = \gamma(f_v) \cdot f_s + \beta(f_v),
\end{equation}
where $f_v$ and $f_s$ are visual and sound features, $\gamma(\cdot)$ and $\beta(\cdot)$ are single linear layers which output scaling and bias on the sound features dependent on visual features.

The output spectrogram mask is obtained after a \texttt{sigmoid} activation on the network output. Then it is thresholded and multiplied with the input spectrogram to get a predicted spectrogram. Finally, an inverse Short Time Fourier Tranform (iSTFT) is applied to obtain the separated sound.

\subsection{Curriculum Learning}
\label{sec:curriculum}
Directly training sound separation on a single class of instruments suffers from overfitting due to the limited number training samples we have for each class. To remedy this drawback, we propose a 3-stage curriculum training by bootstrapping the model with easy tasks for good initializations, so that it converges better on the main tasks. The details are outlined as follows: 

\begin{enumerate}[label=(\roman*)] 
	\item Sound separation on mixture of different instruments. It shares similar settings as Section \ref{sec:separation_different}, where we randomly sample two video shots from the whole training set, mix their sounds as model input for separation; 
	\item Sound separation on mixture of the same kinds of instruments. Initializing from the model weights trained in Step 1, we then only train the model with mixtures from the same instruments, \eg. two videos of cellos;
	\item Sound separation on mixture from the same video. To form the mixture, we sample two different video shots from the same long video. This is the hardest stage as semantic and context cues of those videos can be exactly the same, and the only useful cue is motions.
\end{enumerate}

Note that we will only use this curriculum learning strategy in the same instrument sound separation task due to its challenging nature.

\section{Experiments}

\subsection{Dataset}
We perform vision guided sound separation tasks on the mixture of two video datasets: MUSIC~\cite{Zhao_2018_ECCV} and URMP~\cite{li2019creating}. MUSIC is an unlabeled video dataset of instrument solos and duets by keyword query from Youtube; URMP is a small scale high quality multi-instrument video dataset recorded in studio.

To prevent the models from overfitting, we enlarge the MUSIC~\cite{Zhao_2018_ECCV} dataset by collecting a larger number of musical instrument categories from web videos. Apart from the 11 instrument categories defined in MUSIC dataset: accordion, acoustic guitar, cello, clarinet, erhu, flute, saxophone, trumpet, tuba, violin and xylophone, we include another 10 common instrument categories: bagpipe, banjo, bassoon, congas, drum, electric bass, guzheng, piano, pipa and ukulele. We follow the procedure of~\cite{Zhao_2018_ECCV} to collect the videos. Specifically, we construct a keyword with both instrument name with an additional ``cover''  and use it to retrieve videos from YouTube. We name the resulting dataset MUSIC-21, it contains 1365 untrimmed videos of musical solos and duets, where we split them into a training set of 1065 videos and a test set of 300 videos.

As our trajectory-based representation is sensitive to shot changes, we pre-process the raw videos into video shots, so that our training samples do not cross shot boundaries. Concretely, we densely sample the video frames and calculate the color histogram change of the adjacent frames over time, then we use a double thresholding approach \cite{canny1986computational} to find shot boundaries. After the processing, we obtain 5861 video shots in total.

\subsection{Sound Separation for Different Instruments}
\label{sec:separation_different}

To verify the effective of the learning motion representation for sound separation, we first evaluate the model performances in the task of separating sounds from different kinds of instruments, which has been explored in other works~\cite{virtanen2007monaural,chandna2017monoaural,Zhao_2018_ECCV,gao2018object-sounds}.

\subsubsection{Experiment Configurations}
During training, we randomly take 3-second video clips from the dataset, and then sample RGB frames at 8 FPS to get 24 frames, and sample audios at 11 kHz.

The motion network takes 24 RGB frames as input. The flow network (PWC-Net) in it estimates 23 dense optical flow fields; the trajectory estimator further extracts trajectories with length of 23; and then the trajectory features are extracted by I3D. The output feature maps are of size $T \times H \times W \times K_m$.

The appearance network takes the first frame of the clip, and outputs appearance feature of size $1 \times H \times W \times K_a$. This feature is fused with the trajectory features through the fusion module, and after spatial pooling, we obtain the appearance feature of size $T \times K_v$. 

The sound separation network takes a 3-second mixed audio clip as input, and transforms it into spectrogram by Short Time Fourier Transform (STFT) with frame size of 1022 and hop size of 172. The spectrogram is then fed into a U-Net with 6 convolution and 6 deconvolution layers. 
In the middle of the sound separation network, visual features are aligned with the sound features, and the FiLM module modulates the sound features conditioned on visual features.
The U-Net outputs a binary mask after \texttt{sigmoid} activation and thresholding. To obtain the final separated audio waveforms, iSTFT with the same parameters as the STFT is applied.

We use SGD optimizer with 0.9 momentum to train the our model. The Sound Separation Network and the fusion module use a learning rate of 1e-3; the Motion Network and Appearance Network use a learning rate of 1e-4, as they take pretrained ResNet and I3D on ImageNet and pretrained PWC-Net on MPI Sintel.

\begin{table}[t]
	\begin{center}
		\begin{tabular}{l | c  c c c  c c}
			\specialrule{.2em}{.1em}{.1em}
			Method & SDR &  SIR & SAR \\ \hline
			
			NMF~\cite{virtanen2007monaural} & 2.78   & 6.70  &  9.21  \\
			Deep Separation~\cite{chandna2017monoaural}  & 4.75 & 7.00 & 10.82  \\
			MIML~\cite{gao2018object-sounds} & 4.25 & 6.23 & 11.10 \\
			Sound of Pixels~\cite{Zhao_2018_ECCV} & 7.52 & 13.01 & 11.53 \\ \hline
			Ours, RGB single frame &  7.04  & 12.10  & 11.05 \\ 
			Ours, RGB multi-frame  & 7.67  & 14.81 & 11.24 \\ 
			Ours, RGB+Flow   & 8.05 & 14.73 & 12.65 \\
			Ours, RGB+Trajectory & \textbf{8.31} & \textbf{14.82} & \textbf{13.11} \\ 
			
			\specialrule{.1em}{.05em}{.05em}
		\end{tabular}
	\end{center}
	\caption{Sound source separation performance ($N=2$ mixture) of baselines and our model with different input modalities. Compared to Sound of Pixels, our models with temporal information perform better in sound separation.}
	\label{tab:eval_sota}
\end{table}

\begin{table}[t]
	\begin{center}
		\begin{tabular}{l | c | c c c }
			\specialrule{.2em}{.1em}{.1em}
			
			N & Method & SDR &  SIR & SAR \\ \hline
			\multirow{3}{*}{3} & NMF~\cite{virtanen2007monaural} &  2.01   & 2.08    &  9.36 \\
			& Sound of Pixels~\cite{Zhao_2018_ECCV} &  3.65   &  8.77   &  8.48  \\
			& Ours, RGB+Trajectory   &  4.87   & 9.48    &  9.24 \\ \hline
			\multirow{3}{*}{4} & NMF~\cite{virtanen2007monaural} &  0.93   & -1.01   &  9.01 \\
			& Sound of Pixels~\cite{Zhao_2018_ECCV} &  1.21   & 6.58    &  4.19 \\
			& Ours, RGB+Trajectory   &  3.05   & 8.50  &  7.45 \\
			
			\specialrule{.1em}{.05em}{.05em}
		\end{tabular}
	\end{center}
	\caption{Sound separation performances with $N=3,4$ mixtures. We compare our model against Sound of Pixels to show the advantage of motion features. Our model consistently improves separation metrics and outperforms in highly mixed cases.}
	\label{tab:eval_n}
\end{table}

\subsubsection{Results}
We evaluate the sound separation performance of our model with different variants. \textbf{RGB+Trajectory} is our full model as described in \ref{sec:arch}; \textbf{RGB+Flow} is the full model without the tracking module, so the motion feature is extracted from optical flow; \textbf{RGB multi-frame} further removes the flow network, so motion feature directly comes from RGB frame sequence; \textbf{RGB single frame} is a model without motion network, visual feature comes from appearance network only.

At the same time, we re-implement 4 models to compare against. \textbf{NMF}~\cite{virtanen2007monaural} is a classical approach based on matrix factorization, it uses ground truth labels for training; \textbf{Deep Separation}~\cite{chandna2017monoaural} is a CNN-variant supervised learning model, it also takes ground truth labels for training; \textbf{MIML} \cite{gao2018object-sounds} is a model that combines NMF decomposition and multi-instance multi-label learning; \textbf{Sound of Pixels}~\cite{Zhao_2018_ECCV} is a recently proposed self-supervised model which takes both sounds and video frames for source separation.

For fair comparisons, all the models are trained and tested with 3-second audios mixed from $N=2$ input audios, and models dependent on vision take in 24 video frames. Model performances are evaluated on a validation set with 256 pairs of sound mixtures.
We use the following metrics from the open-source \texttt{mir\_eval}~\cite{raffel2014mir_eval} library to quantify performance: Signal-to-Distortion Ratio (SDR), Signal-to-Interference Ratio (SIR), and Signal-to-Artifact Ratio (SAR). Their units are in dB.


\begin{figure*}[t]
	\centering
	\includegraphics[width=1.\linewidth]{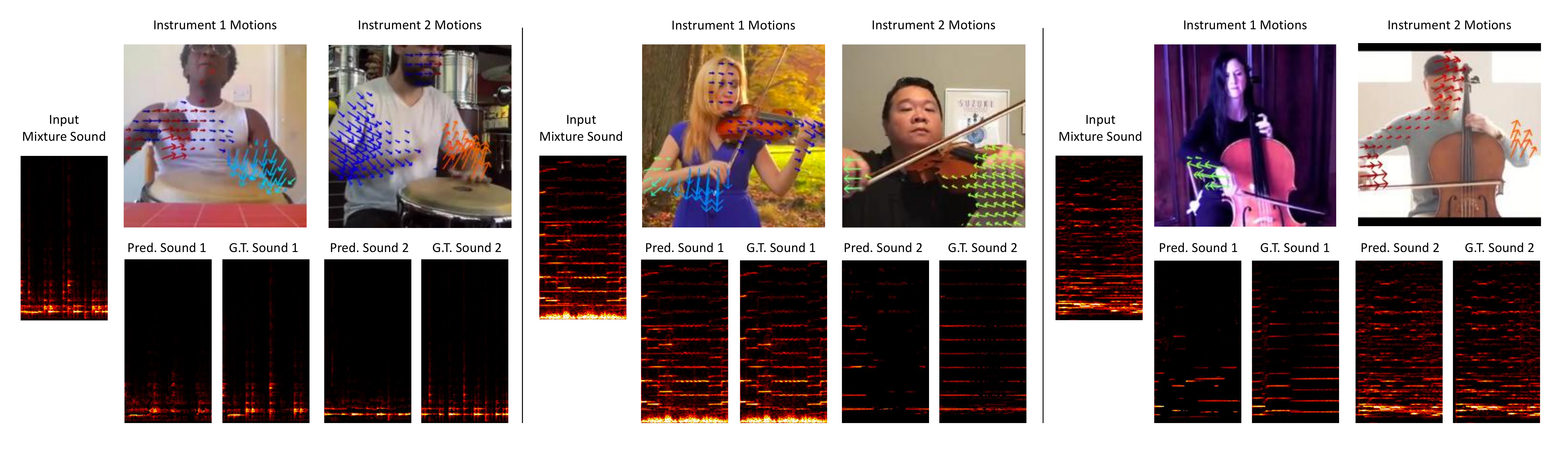}
	\caption{Results of sound separation on the same kinds of instruments. Our model can capture the motion information in videos to separate the sound. This visualization is only performed for quantitative model evaluation.}
	\label{fig:result_separation_same}
\end{figure*}


Quantitative results are reported in Table \ref{tab:eval_sota}. We observe that previous methods achieves reasonable performance in sound separation even though only appearance information is used~\cite{Zhao_2018_ECCV}. It shows that appearance based models are already strong baselines for this task.
In comparison, our RGB multi-frame, RGB+Flow and RGB+Trajectory models outperform all baseline methods, showing the effectiveness of encoding motion cues in the task of audio-visual source separation. And among them, RGB+Trajectory is best, and outperforms state-of-the-art Sound of Pixels model by $\approx0.8 \text{dB}$. It demonstrates that among these motion representations, trajectories has the strongest correlation with sound.

We further experiment on the task of separating larger number of sound mixtures, where $N=3,4$. Results are reported in Table~\ref{tab:eval_n}. We observe that our best model outperforms Sound of Pixels by a larger margin in these highly mixed cases, $\approx1.2 \text{dB}$ at $N=3$, and $\approx1.8 \text{dB}$ at $N=4$.

\subsection{Sound Separation for the Same Instruments}
\label{sec:separation_same}

In this section, we evaluate the model performance in separating sounds from instruments of the same kind, which has rarely been explored before.

\subsubsection{Experiment Configurations}
To evaluate the performances of the our models, we select 5 kinds of musical instruments whose sounds are closely related to motions: violin, cello, congas, erhu and xylophone. All the training settings are similar to Section~\ref{sec:separation_different} except that we use curriculum learning strategy which is mentioned in Section \ref{sec:curriculum}.

\subsubsection{Results}
\label{sec:separation_same_results}
First we evaluate the effectiveness of our proposed curriculum learning strategy. With a fixed validation set, we compare \textbf{Single Stage} strategy, which is directly trained on mixtures of the same instruments, with our 3-stage training strategy. In \textbf{Curriculum Stage 1}, model is trained to separate sound mixtures of instruments of different categories; in \textbf{Curriculum Stage 2}, the task is to separate sound mixtures from the same kinds of instruments; in \textbf{Curriculum Stage 3}, the goal is to separate sound mixtures of different clips from the same long video. Results of our final model on the validation set are shown in Figure \ref{fig:result_separation_same}.

Results in Table~\ref{tab:eval_curriculum} show that curriculum learning greatly improves the performance: it outperforms the \textbf{Single Stage} model in the \textbf{Curriculum Stage 1}, and further improves with the second and third stages. The total improvement in SDR is $\approx4 \text{dB}$.

\begin{table}[t]
	\begin{center}
		\begin{tabular}{l|ccc}
			\specialrule{.2em}{.1em}{.1em}
			Schedule & SDR & SIR & SAR \\ \hline
			Single Stage & 1.91   & 5.73  & 8.83 \\ \hline
			Curriculum Stage 1 &  3.14   & 7.52  & 13.06 \\
			Curriculum Stage 2 &  5.72   & 13.89  & 11.92 \\
			Curriculum Stage 3 &  5.93   & 14.41  & 12.08 \\
			\specialrule{.1em}{.05em}{.05em}
		\end{tabular}
	\end{center}
	\caption{Performance improvement with the proposed curriculum learning schedule.}
	\label{tab:eval_curriculum}
\end{table}

Then we compare the performance of our model with Sound of Pixels model on the same instrument separation task. To make fair comparisons, Sound of Pixels model is trained with the same curriculum. Results on SDR metric are reported in Table~\ref{tab:eval_same}. We can see that Sound of Pixels model gives much inferior results comparing to our model, the gap is $>3 \text{dB}$.

\begin{figure}[t]
	\centering
	\includegraphics[width=1.0\linewidth]{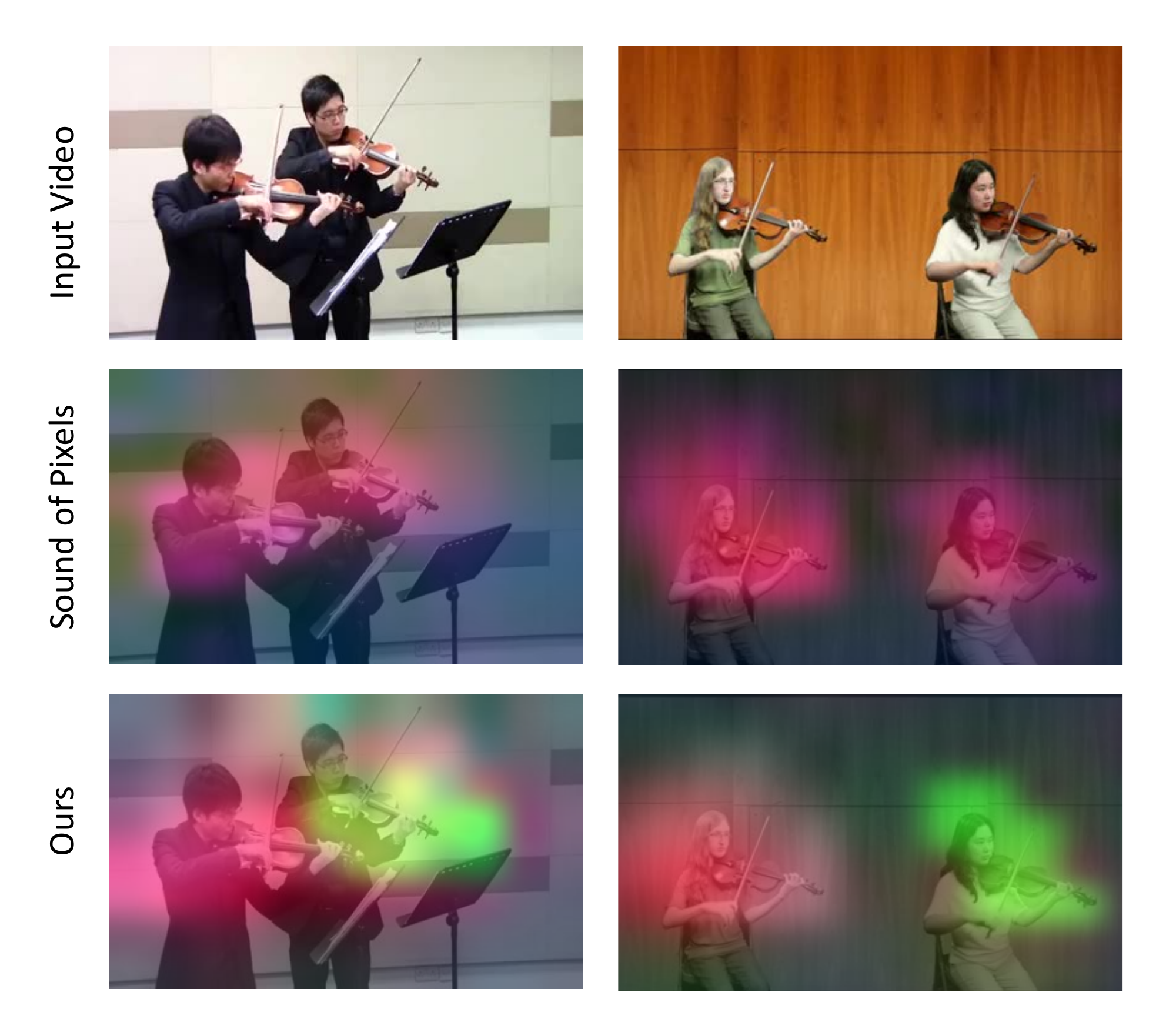}
	\caption{Pixel-level sound embedding results. To visualize the pixel-level sound separation results, we project sound features into a low dimensional space, and visualize them in RGB space. Different colors mean different sounds. Our model can tell the difference from duets of the same instruments, while Sound of Pixels model cannot.}
	\label{fig:result_embedding}
\end{figure}

Qualitative comparisons are presented in Figure \ref{fig:result_embedding}, where we show pixel-level sound embeddings. To recover sounds spatially, we remove the spatial pooling operation in the fusion module in Figure~\ref{fig:model_fusion} at test time, and then feed the visual feature at each spatial location to the Sound Separation Network. Therefore, we are able to get $H\times W$ number of separated sound components. We project those sound features (vectorized spectrogram values) into a 3 dimensional space using PCA, and visualize them in color. Different colors in the heatmaps refer to different sounds. We show that our model can tell the difference from duets of the same instruments, while Sound of Pixels model cannot.

\begin{table}[t]
	\begin{center}
		\begin{tabular}{l|ccccc}
			\specialrule{.2em}{.1em}{.1em}
			Instrument & Sound of Pixels & Ours\\ \hline
			violin    & 1.95  & 6.33 \\ \hline
			cello     & 2.62  & 5.48  \\ \hline
			congas    & 2.90  & 5.21 \\ \hline
			erhu      & 1.67  & 6.13 \\ \hline
			xylophone & 3.56  & 6.50 \\
			\specialrule{.1em}{.05em}{.05em}
		\end{tabular}
	\end{center}
	\caption{Sound source separation performance on duets of the same instruments. We show the SDR metric on each instrument. Our approach is consistently better than previous works.}
	
	\label{tab:eval_same}
\end{table}


\subsubsection{Human Evaluation}

Since the popular metrics (\eg SDR, SIR and SAR) for sound separation might not reflect the actual perceptual quality of the sound separation results, we further compare the performances of these two methods on Amazon Mechanical Turk (AMT) with subjective human evaluations.

Concretely, we collected 100 testing videos from each instrument, and got separation results of the Sound of Pixels baseline~\cite{Zhao_2018_ECCV} and our best model. We also provide the ground truth results for references. To avoid shortcut, we randomly shuffle the orders of two models and ask the following question: \texttt{Which sound separation result is closer to the ground truth?} The workers are asked to choose one of the best sound separation results. We assign 3 independent AMT workers for each job.

Results are shown in Table~\ref{tab:eval_amt}, our proposed motion-based model consistently outperforms the Sound of Pixels systems for all the five instruments. We see the reasons lie in two folds: (1) motion information is crucial for the sound separation of the same instruments; (2) the Sound of Pixels model cannot capture motion cues effectively, while our model is better by design.

\begin{table}[t]
	\begin{center}
		\begin{tabular}{l|ccccc}
			\specialrule{.2em}{.1em}{.1em}
			Instrument & Sound of Pixels & Ours\\ \hline
			violin    & 38.75\%  & 61.25\% \\ \hline
			cello     & 39.21\%  & 60.79\%  \\ \hline
			congas    & 35.42\%  & 64.58\% \\ \hline
			erhu      & 44.59\%  & 55.41\% \\ \hline
			xylophone & 35.56\%  & 64.44\% \\
			\specialrule{.1em}{.05em}{.05em}
		\end{tabular}
	\end{center}
	\caption{Human evaluation result for the sound source separation on mixture of the same instruments.}
	
	\label{tab:eval_amt}

\end{table}


\begin{figure}[t]
	\centering
	\includegraphics[width=1.0\linewidth]{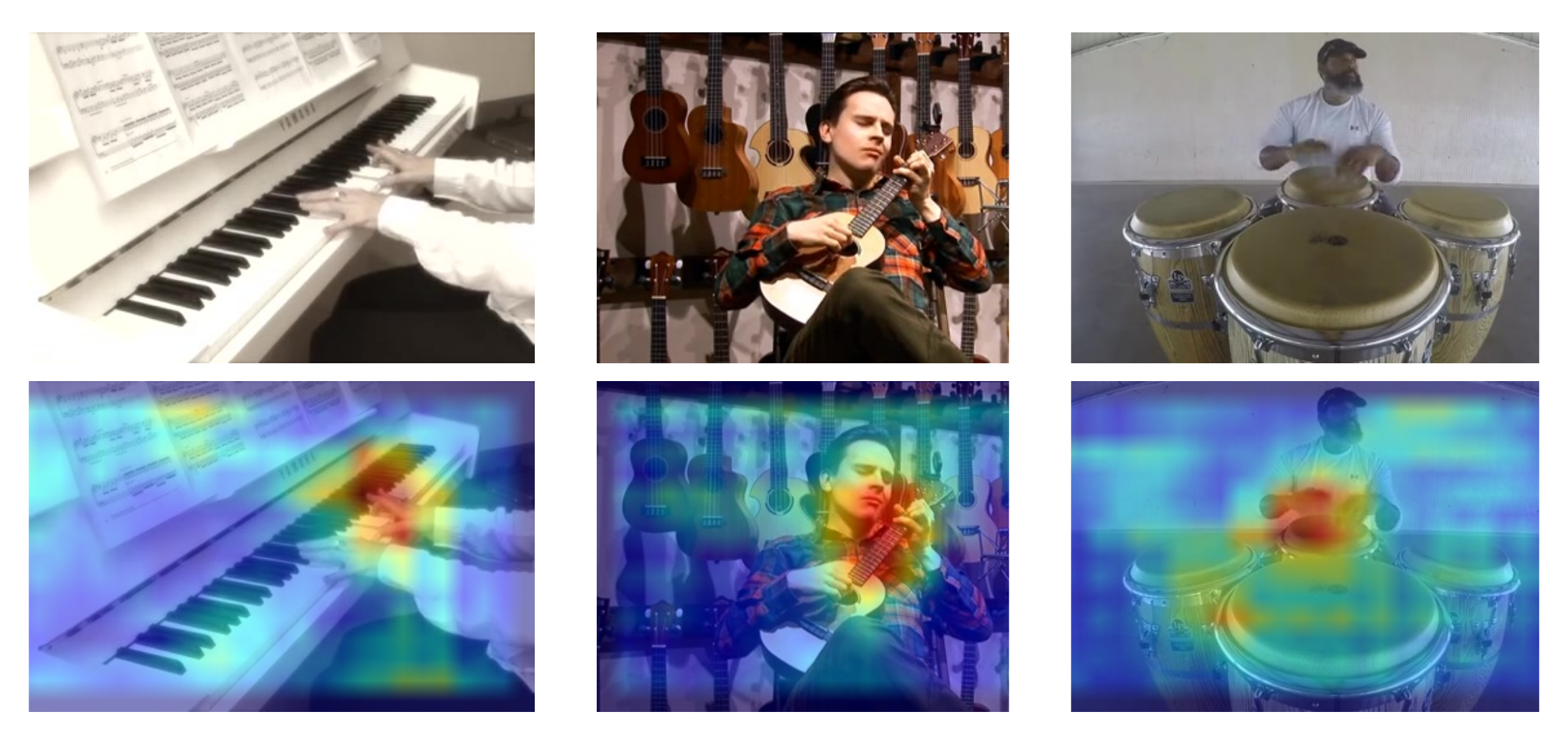}
	\caption{Sounding object localization. Overlaid heatmaps show the predicted sound volume at each pixel location. The model tends to predict the instrument parts where people are interacting with. Silent instruments such as the guitars on the wall are not detected as sounding objects.}
	\label{fig:result_localization}
\end{figure}

\subsection{Sounding object localization}
As a further analysis, we explore the sounding object localization capability of our best model. We recover the sounds spatially similar to what we did in Section~\ref{sec:separation_same_results}. And then we calculate the sound volume at each spatial location, and display them in heatmaps, as shown in Figure~\ref{fig:result_localization}. We observe that (1) the model gives roughly correct predictions on the sounding object locations, but does not cover the whole instruments. Interestingly, it focuses on the parts where humans are interacting with; (2) Our model correctly predicts silent instruments, \textit{e.g.} guitars on the wall, it demonstrates that sounding object localization is not only based on visual appearance, but also on audio input.

\section{Conclusion}
\label{sec:conclusion}

In this paper, we propose that motions are important cues in audio-visual tasks, and design a system that captures visual motions with deep dense trajectories (DDT) to separate sounds. Extensive evaluations show that, compared to previous appearance based models, we are able to perform audio-visual source separation of different instruments more robustly; we can also separate sounds of the same kind of instruments through curriculum learning, which seems impossible for the purely appearance based approaches.

{\small
\bibliographystyle{ieee}
\bibliography{egbib}
}

\end{document}